\def\year{2022}\relax

\documentclass[letterpaper]{article}
\usepackage{aaai22}
\usepackage{times}
\usepackage{helvet}
\usepackage{courier}
\usepackage[hyphens]{url}  
\usepackage{natbib}  
\usepackage{caption} 
\usepackage{amsmath,amssymb,amsfonts}
\usepackage{subcaption}
\usepackage{booktabs}
\usepackage{algorithmic}
\usepackage[ruled,vlined, linesnumbered]{algorithm2e}
\SetKwComment{Comment}{$\triangleright$\ }{}
\usepackage{graphicx}
\usepackage{textcomp}
\usepackage{xcolor}
\usepackage{comment}
\DeclareCaptionStyle{ruled}{labelfont=normalfont,labelsep=colon,strut=off} 

\begin{document}
%
\title{Clustering with UMAP: Why and How Connectivity Matters}

\author {
    Ayush Dalmia, Suzanna Sia
}
\affiliations {
    Department of Computer Science,
    Johns Hopkins University\\
    adalmia96@gmail.com, ssia1@jhu.edu
}



\maketitle

\begin{abstract}
\begin{quote}
Topology based dimensionality reduction methods such as t-SNE and UMAP have strong mathematical foundations and are based on the intuition that the topology in low dimensions should be close to that of high dimensions. Given that the initial structure is a precursor to the success of the algorithm, this naturally raises the question: What makes a “good” topological structure for dimensionality reduction? In this paper which focuses on UMAP, we study the effects of node connectivity (k-Nearest Neighbors vs \textit{mutual} k-Nearest Neighbors) and relative neighborhood (Adjacent via Path Neighbors) on dimensionality reduction. We explore these concepts through extensive ablation studies on 4 standard image and text datasets; MNIST, FMNIST, 20NG, AG, reducing to 2 and 64 dimensions. Our findings indicate that a more refined notion of connectivity (\textit{mutual} k-Nearest Neighbors with minimum spanning tree) together with a flexible method of constructing the local neighborhood (Path Neighbors), can achieve a much better representation than default UMAP, as measured by downstream clustering performance.
\end{quote}
\end{abstract}

\section{Introduction}
Dimension reduction techniques have become a standard tool in both machine learning and data analysis. 
Topology based dimensionality reduction techniques such as $t$-SNE \cite{van2008visualizing} and Uniform Manifold Approximation and Projection (UMAP; \citet{mcinnes2018umap}) have been increasing in popularity due to their effectiveness in data visualisation and finding better representations for downstream tasks such as classification and clustering. While various topological dimensionality reduction methods differ in mathematical details for optimizing the low-dimensional manifold, these methods typically start with an initial graph structure in high dimensions. 
This naturally raises the question: \textbf{What makes a “good” topological structure for dimensionality reduction?} We explore different notions of connectivity for the initial graph structure and evaluate the performance of the resulting low-dimensional representation on a downstream clustering task using Normalised Mutual Information (NMI).

The most general initialization can be straightforwardly obtained by a weighted $k$ nearest neighbor (k-NN) graph. 
However, the k-NN graph is relatively susceptible to the “curse of dimensionality” and the associated \textit{distance concentration effect}, where distances are similar in high dimensions, as well as the \textit{hub effect}, where certain points become highly influential when highly connected. This skews the local representation of high dimensional data, deteriorating its performance for various similarity-based machine learning tasks such as classification \cite{TOMASEV2015157, Dinu2015ImprovingZL}, clustering \cite{Tomasev2014} and most notably dimension reduction \cite{Feldbauer2018ACE}. 

In this paper, we focus on UMAP, which 
has advantages over other topology based dimension reduction techniques such as PCA, Isomap \cite{balasubramanian2002isomap}, and t-SNE for visualization quality, preserving a data’s global structure (inter-class distances) and its local structure (intra-class distances), and having superior runtime performance. 
UMAP has been increasingly used in various real world applications such as vision \cite{Mebarka2020, VERMEULEN2021119547}; NLP \cite{kayal-2021-unsupervised, rother-etal-2020-cmce}; population genetics \cite{diaz2020review, Becht2019DimensionalityRF} with competitive performance on unsupervised clustering and outlier detection among others. 

In Section 3, we propose a refinement in the graph construction stage of the UMAP algorithm that uses a \textit{mutual} k-NN graph instead of k-NN, to reduce the undesired \textit{distance concentration} and \textit{hub effects}. A consequence of applying \textit{mutual} k-NN graph is that it can results in isolated components that can adversely affect the initialization of the lower dimensional vectors. Therefore in Section 3.2 we explore different methods such as linking adjacent neighbors, and adopting the edges from a minimum spanning tree (MST) to link the isolated components suggested by prior research. The resulting NN graph can be refined to obtain the NN for each point, by using the Shortest Path Distance so that non-adjacent neighbors (which we term Path Neighbors) can also be considered as the local neighborhood.

In Section 4, we conduct experiments on 4 standard image and text datasets, and visualise the resulting representation after applying UMAP, which suggest how each variant of Nearest Neighbor affects the low dimensional representation.  The contributions of this paper are:

\begin{itemize}

\item using a \textit{mutual} k-NN graph rather than the original k-NN graph improves the separation between similar classes for both image and text datasets.


\item having a minimally connected graph by adding the minimum edges from the MST (MST-min) is best if we have access to flexible methods of constructing the local neighborhood (e.g via MST-min with Shortest Path distance neighbors (Path Neighbors)).
\item we show that simple graph processing can result in relative gains of 5-18\% on NMI consistently across all datasets, and analyse the effect of graph processing on downstream representations.


\end{itemize}

\section{Background}

\subsection{Understanding UMAP}
\label{sec:umap}
UMAP first constructs a high dimensional graph representation of the data (graph construction phase), which is used to represent the “fuzzy simplicial complex”. Next it optimizes low-dimensional vectors to be as similar as possible to the graph representation in higher dimensions, by minimizing the cross-entropy of two fuzzy sets with the same underlying elements (datapoints). 



\subsubsection{Graph Construction}

Our work is concerned with the graph construction phase of UMAP. In default UMAP, a weighted k-NN graph is constructed from the data where each vertex represents a datapoint, and edge weights represents the likelihood that two points are connected \cite{mcinnes2018umap}. Formally, let $X = \{x_1, \cdots, x_N \}$ be a dataset with $N$ points, and $x\in \mathbb{R}^M$. For each datapoint $x_i$, the set of points in its local neighborhood is computed using a distance metric $d: X \times X \rightarrow \mathbb{R}$. 
In default UMAP, the neighborhood is found using the nearest neighbor descent algorithm \cite{dong2011efficient}. Then for each $x_i$, we have $\rho_i$ which reflects a connectivity constraint that data points are assumed to be locally connected. Here $\rho_i = \mathrm{min}\{d(x_i, x_j) | d(x_i, x_j) > 0, x_j\in X\}$. This ensures that $x_i$ connects to at least one other datapoint.
We define a weighted Graph $G\!=\!(V, E, w)$ where the vertices $V$ are simply the set of datapoints $X$ and $w$ is the set of weights corresponding to $E$. $G$ represents the “fuzzy simplicial complex”. Details on how to construct $G$ can be found in \citet{mcinnes2018umap}. 



The central focus of this paper is to demonstrate how refining the set of directed edges $E$ which reflects the connectivity of the graph in high-dimensional representations, affects the subsequent low-dimensional representations. The set of directed edges is $E = \{(x_i, x_j) | i \leq N, x_j \in \mathcal{M}_i \}$, and we define $\mathcal{M}_i$ 
to be the local neighborhood of points for $x_i$ that collectively determine the final set of directed edges $E$. In this work, our starting point for $\mathcal{M}_i$ is the \textit{mutual} nearest neighbors and we deal with the inherent asymmetry of this graph which could result in data points being completely isolated (Section 3.1).


\subsubsection{Optimizing lower dimensional representation}

Details about this optimization are not critical to understanding our proposed method, and we present this for completeness. Given $G$, the data is projected into lower dimensions via a force-directed graph layout algorithm. To initialize the lower dimensional vectors, UMAP uses the eigenvectors of a normalized graph Laplacian of the  “fuzzy simplicial complex” affinity matrix, also known as spectral vectors. For theoretical foundations and  mathematical details, we refer interested readers to \cite{mcinnes2018umap}.

\begin{figure*}[t!]
  \centering
    \begin{subfigure}{0.3\textwidth}
    \includegraphics[width=\textwidth]{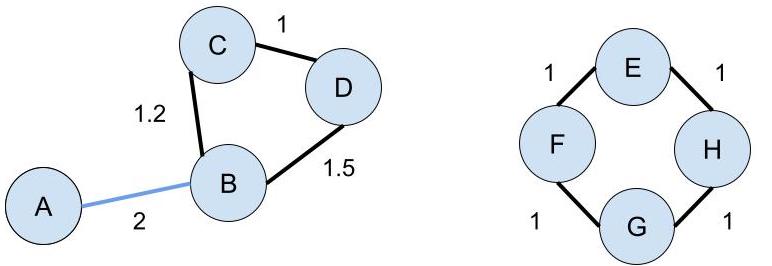}
    \subcaption{Edge(blue) added by NN}
    \end{subfigure}\hspace{5mm}
    \begin{subfigure}{0.3\textwidth}    \includegraphics[width=\textwidth]{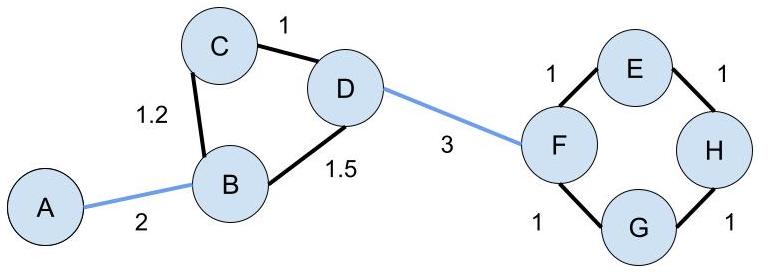}
    \subcaption{Edge(blue) added by MST-min}
    \end{subfigure}\hspace{5mm}
    \begin{subfigure}{0.3\textwidth}
    \includegraphics[width=\textwidth]{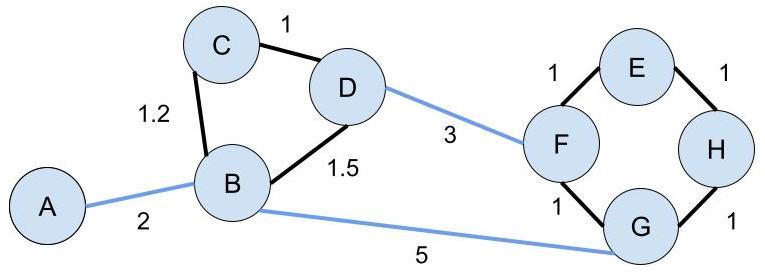}
    \subcaption{Edge(blue) added by MST-all}
    \end{subfigure}
  \caption{\small Visual comparisons between all the methods used in Section 3.1. Black edges  represent edges from the \textit{mutual} k-NN graph. Blue edges represent the edges added using each connectivity method: connecting isolated nodes by nearest neighbor (NN), connecting components with the minimum edges of a minimum spanning tree (MST-min), and with all the edges from the minimum spanning tree (MST-all).}
  \label{first-figure-with-subfigures}
\end{figure*}

\subsection{Mutual k-Nearest Neighbors}
The default choice of $\mathcal{M}_i$ are the nearest neighbors of $x_i$. However, previous research has shown that a weighted k-NN graph captures noisy relationships, and may not be an accurate representation of the underlying local structure for a high dimensional dataset due to the “curse of dimensionality” \cite{liu2012noisy, Radovanovi2010HubsIS}. Instead we utilize a weighted \textit{mutual} $k$-nearest neighbor graph (weighted \textit{mutual} k-NN graph) to represent the underlying structure of our data. A \textit{mutual} k-NN graph is defined as a graph that only has an undirected edge between two vertices $x_i$ and $x_j$ if $x_i$ is in $k$-nearest neighbors of $x_j$ and $x_j$ is in the $k$-nearest neighbors of $x_i$. Therefore our local neighborhood is $\mathcal{M}_i = \{x_j | x_j \in knn(x_i), x_i \in knn(x_j) \}$.

A \textit{mutual} k-NN graph can hence be interpreted as a subgraph of the original k-NN graph. \textit{Mutual} k-NN graphs have been shown to contain many desirable properties in comparison to its k-NN counterpart when combating the “curse of dimensionality”. One such property is that edges in \textit{mutual} k-NN graphs have been shown to be a stronger indicator of similarity than the unidirectional nearest neighborhood relationship, alleviating the \textit{distance concentration effect} \cite{Jegou2010}. \textit{Mutual} k-NN graphs have also been shown to be less prone to the \textit{hub effect} than the regular k-NN graph, since each vertex in a \textit{mutual} k-NN graph  is guaranteed to be at most k-degrees \cite{ozaki-etal-2011-using}. They have also shown strong performance when used in both unsupervised clustering \cite{Maier2009, Sardana2014, Abbas2012} and semi-supervised graph based classifications algorithms \cite{ozaki-etal-2011-using, de2013influence}. In Section 3, we describe how we utilize the \textit{mutual} k-NN graph to create our high dimensional graph representation.



\section{Methodology}
\label{sec:methodology}

In order to obtain the fuzzy simplicial complex for UMAP, our starting point is a weighted \textit{mutual} k-NN graph as motivated in Section 2.2. However this graph may have completely isolated vertices and disconnected components, violating UMAP's assumption that the manifold is locally connected (Section 2.1). In addition, having many components is undesirable for the initialization of the lower dimensional spectral vectors \cite{mcinnes2018umap}. Therefore, we consider different connectivity methods such as using the minimum spanning tree (MST) to add edges, which we elaborate in Section 3.1. 

Next, we consider that using a vertex's adjacent neighborhood for $\mathcal{M}_i$ may not pose an adequate representation as this excludes all points that are more than 1 hop away. Hence we additionally refine $\mathcal{M}_i$ using the ``new" neighborhood calculated using Shortest Path Distance to other nodes on the connected graph. This is elaborated in Section 3.2. Our method can thus be viewed as first having a loosely connected (\textit{mutual} k-NN) graph, followed by increasing connectivity using MST variants, and finally with a refined $\mathcal{M}_i$ to define neighbors that are more than one hop away.

\subsection{Increasing Connectivity for \textit{mutual} k-NN}
Since a \textit{mutual} k-NN graph only retains a subset of the edges from the original k-NN graph(Section 2.2), the resulting \textit{mutual} k-NN graph often contains disconnected components and potential isolated vertices. Isolated vertices violate one of UMAP’s primary assumptions that the underlying manifold is locally connected, and disconnected components negatively impact the spectral initialization \cite{mcinnes2018umap}. Spectral initialization is crucial for preserving the global structure of the high dimensional data in the low dimensional vectors \cite{Kobak2021}.

Naively, we could increase $k$, the number of nearest neighbors used for calculating the \textit{mutual} k-NN graph, to obtain a more connected graphical representation. However increasing $k$ does not guarantee that the resulting graph has no isolated vertices or fewer disconnected components. 
We therefore consider the following methods for increasing the Connectivity of the \textit{mutual} k-NN graph, presented in order of increasing connectivity.

\begin{enumerate}
\item \textbf{NN}: 
Add an undirected edge between each isolated vertex and its original nearest neighbor \cite{de2013influence}. 
\item \textbf{MST-min}: To achieve a connected graph, add the minimum number of edges from a maximum spanning tree to the mutual-kNN graph that has been weighted with similarity-based metrics\cite{ozaki-etal-2011-using}. We adapt this by calculating the minimum spanning tree for distances (Algorithm 1).
\item \textbf{MST-all}: Adding all the edges of the MST.
\end{enumerate}

We call this graph $G'$, the locally connected \textit{mutual} k-NN graph, and run experiments with $G'$ in Section 4.

\subsection{Finding New Local Neighborhood $\mathcal{M}_i$}
Next we wish to obtain a “fuzzy simplicial complex” representation as described in Section 2.1, from the connected \textit{mutual} k-NN graph, $G'$ (Section 2.2). To achieve this, we first need to obtain the new local neighborhood, $\mathcal{M}_i$ for each $x_i \in X$ using $G'$. The most straightforward way is to use adjacent vertices in $G'$. 
However, this excludes all points that are more than 1 hop away, even if they are relatively close.

Instead, we compute $G$ from $G'$ by refining the local neighborhood $\mathcal{M}_i$ using the Shortest Path Distance between any two nodes using Dijkstra's Algorithm (See Fig. 2 for example) to other nodes on $G'$ (Algorithm 2). 
This shortest path distance can be considered a new distance metric as it directly aligns with UMAP’s definition of an extended-pseudo-metric space. That is, we replace $d(x_i, x_j’)$ with $d_{\mathrm{path}}(x_i, x_j)$  in the graph construction phase (Section 2.1).

\begin{figure}[h]
  \centering
  \includegraphics[width=0.5\linewidth, height = 3cm]{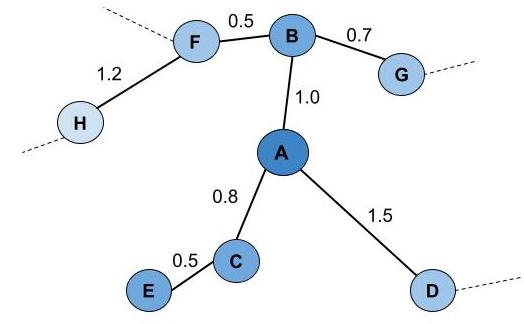}
  \caption{\small Example of the local neighborhood we may encounter. If we are trying to find the neighborhood for A, then we would only select nodes B, C, D with Adjacent Neighbors. However this excludes closer nodes such as E and F so by using Path Neighbor, we can include E and F in the neighborhood of A.}
  \label{fig:local_neigh}
\end{figure}

\begin{algorithm}[t!]
\small
\SetAlgoLined
\KwIn{ The mutual k-NN graph $mkNN$, kNN Graph $kNN$ with weights as the distance between points} 
\KwOut{Connected $mkNN$ graph $G'$ }
 Initialize $G'$ as a copy of $mkNN$\\
$MST \leftarrow$ minimum weight spanning tree of $kNN$\\
$sort\_e \leftarrow$ sort MST edges by ascending weight order\\
\ForEach{$ \text{edge } e \in sort\_e$} {
        \If{$e \text{ connects two components in } G'$} {
            Add undirected edge $e$ to $G'$
        }
}

\Return $G'$
 \caption{\small Connect Disconnecting Components in \textit{mutual} k-NN Graph with MST-min }
\end{algorithm}

\begin{algorithm}
\small 
\SetAlgoLined
\KwIn{Connected mutual k-NN graph $G'$, number of new nearest neighbors to search for $k_{new}$}
\KwOut{Dictionary that returns the new neighborhood for each point $\mathcal{M}$, dictionary that returns the distance between $x$ and its nearest Path Neighbors $\mathcal{M}_{dists}$ }
Initialize $\mathcal{M}$ and $\mathcal{M}_{dists}$ 

\ForEach{$x_i \in X$} {
    Initialize neighborhood of $x_i$,$\mathcal{M}_i$, and $Q$ as a min heap\\
	Insert each adjacent vertex $v$ to $x$ in $G'$ with the associated distance between $v$ and $x$ to $Q$\\
	\While{$Q \text{ not empty \& } |\mathcal{M}_i| < k_{new}$}{
	    Extract minimum distance vertex $y$  with distance $y_{path\_dist}$ from $Q$\\
		\If{$y \notin \mathcal{M}_i$}{
			Append $y$ to $\mathcal{M}_i$ and $y_{path\_dist}$ to $\mathcal{M}_{dists}(x)$\\
			\ForEach{$v \in$ \text{ Adj. vertices of } $y$} {
				Insert $v$ into $Q$ with a distance of $y_{path\_dist}$ + distance between y and v
			}
		}
    }
}

\Return  $\mathcal{M}$ and  $\mathcal{M}_{distances}$
\caption{\small Find new local neighborhood $\mathcal{M}_i$, with Path Neighbor using Djikstra's}

\end{algorithm}

\subsection{Computational Complexity}
The complexity of constructing the \textit{mutual} k-NN graph from the original k-NN graph is $O(k n)$. The additional complexity of connecting disconnected components varies depending on the method chosen. Adding edges between isolated vertices to their nearest neighbors (\textsl{NN}) is $O(n)$ while the MST variants have a known time complexity of $O(E \log( V ))$ = $O(k n \log( n ))$. Since we construct the MST from the original k-NN graph, we know that $E = n k$ and $V = n$. With the MST, we add either all the edges (\textsl{MST-all}), which has an upper bound of $O(n)$  since there are $n - 1$ edges in the MST, or the minimum number of edges (\textsl{MST-min}) until we have one component, which has an upper bound of $O(n \log(n))$  since we need to sort the edges first.

Finally, 
using Path Neighbors on \textit{mutual} k-NN graph has an additional cost since we have to perform a graph search. We use a variant of Djikstra’s algorithm which has a known time complexity of $O(V + E \log( V ))$. 
Running the search for all $V=n$ nodes and $E \leq n k$, we get a final complexity of  $O(n (n + k n \log( n )))$. It is important to note that the worst case for Djikstra’s algorithm assumes one may need to explore the whole graph as it is typically to find the shortest distance between two points. However, since we are using Djikstra’s to find the new closest points, we in practice are more likely to terminate much earlier. This search can be further optimized by parallelizing the individual searches needed to find new nearest neighbors for each point.

\section{Experiments}
\label{sec:experiments}

\subsection{Datasets}
We used two standard image classification datasets, MNIST \cite{lecun-mnisthandwrittendigit-2010} and  Fashion MNIST (FMNIST; \citet{Xiao2017FashionMNISTAN}), and two standard text classification datasets, 20 newsgroup (20NG),\footnote{http://qwone.com/~jason/20Newsgroups/} and AG's News Topic (AG;\citet{Zhang2015CharacterlevelCN}) for evaluation. For the image classification datasets, no preprocessing was applied and each image was flattened to create a 1D vector. For the text classification datasets, we lowercase tokens, remove stopwords, punctuation and digits, and exclude words that appear in less than 5 documents. After preprocessing, we converted each dataset into both Count vectors and TF-IDF vectors. We opted to use the Count based representation of text as an illustration of datasets that have very high-dimensional and sparse representations.\footnote{TF-IDF is a weighting on the sparse representation.} While our method is applicable to word or document vectors, these are heavily dependent on the text encoder and it is more difficult to interpret findings as we do not know what the dense high-dimensional vector space represents in the first place.   \footnote{Code is available at \url{https://github.com/adalmia96/umap-mnn}.}

\subsection{Evaluation and Performance Metrics}


\textbf{Quantative Evaluation}: We evaluate the methods by comparing the clustering performance of the resulting low dimensional vectors generated. We performed KMeans clustering, assigning the number of clusters to be the ground truth number of topics for each dataset. We use the Normalised Mutual Information (NMI), a widely used clustering metric that ranges from 0 to 1 (1 being a perfect score), to evaluate the clusters.\footnote{NMI was primarily used to evaluate clustering quality for UMAP and subsequent works\cite{mcinnes2018umap}} As UMAP is a stochastic algorithm, the NMI scores presented in Table 1 are averaged across low-dimensional vectors from 5 random seeds. 


\textbf{Qualitative Evaluation}: In addition, we consider several desirable properties for dimensionality reduction in qualitative visualisations, \textit{local structure and global structure}, and separation of classes. A dimensionality reduction method which preserves the \textit{local structure} means that intra-class distances are preserved; i.e. the datapoints belonging to the same class are close to each other. A method which preserves the \textit{global structure} means that inter-class distances are preserved; i.e., clusters of similar classes appear closer together and clusters of dissimilar classes appear farther from each other. Finally, the classes should be well separated (not clumped together).



\subsection{Experimental Conditions}
\label{sec:conditions}

\begin{table*}[tb]
\centering

\label{tab:results_nmi}
\tiny
\resizebox{0.78\textwidth}{!}{\begin{tabular}{@{}lll|lll|lll@{}}
\toprule
\multicolumn{3}{l|}{Normalized Mutual Information} & \multicolumn{3}{l|}{Adjacent Neighbors} & \multicolumn{3}{l}{Path Neighbors} \\ \midrule
Dataset & Dim & UMAP & NN & \begin{tabular}[c]{@{}l@{}} MST-min \end{tabular} & \begin{tabular}[c]{@{}l@{}} MST-all \end{tabular} & NN & \begin{tabular}[c]{@{}l@{}} MST-min \end{tabular} & \begin{tabular}[c]{@{}l@{}}MST-all \end{tabular} \\
MNIST & 2 & 0.854 & 0.917 & \bf{0.920} & \bf{0.920} & 0.898 & \bf{0.920} & \bf{0.920} \\
FMNIST & 2 & 0.615 & 0.669 & \bf{0.693} & \bf{0.694} & 0.648 & \bf{0.698} & 0.649 \\
20NG Count & 2 & 0.446 & 0.467 & 0.474 & 0.481 & 0.511 & \bf{0.523} & \bf{0.524} \\
20NG TF-IDF & 2 & 0.478 & 0.461 & 0.456 & 0.461 & \bf{0.521} & \bf{0.526} & \bf{0.522} \\
AG Count & 2 & 0.589 & 0.582 & 0.571 & 0.589 & 0.589 & \bf{0.630} & \bf{0.642} \\
AG TF-IDF & 2 & 0.503 & 0.455 & 0.475 & 0.506 & 0.520 & \bf{0.540} & \bf{0.545} \\

\midrule
MNIST &  64 & 0.862 & 0.915 & \bf{0.919} & \bf{0.920} & 0.910 & \bf{0.919} & \bf{0.918} \\
FMNIST & 64 & 0.626 & 0.685 & \bf{0.703} & \bf{0.703} & 0.667 & \bf{0.698} & 0.679 \\
20NG Count & 64 & 0.487 & 0.525 & 0.529 & 0.535 & \bf{0.560} & \bf{0.563} & \bf{0.565} \\
20NG TF-IDF & 64 & 0.566 & 0.556 & 0.556 & 0.560 & \bf{0.592} & \bf{0.594} & \bf{0.595} \\
AG Count &64 & 0.612 & 0.633 & 0.633 & 0.638 & \bf{0.660} & \bf{0.660} & \bf{0.666} \\
AG TF-IDF & 64 & 0.575 & 0.548 & 0.548 & 0.558 & 0.593 & \bf{0.612} & \bf{0.615} \\ 
\midrule
\end{tabular}}
\caption{\small NMI Results for clustering each of the vectors generated using each method described in Section 4.3 with KMeans. For all datasets, using a \textit{mutual} k-NN representation with one of the MST variants and combined with Path Neighbors provided the best NMI results.}
\end{table*}

Our starting point for all connectivity methods is the \textit{mutual} k-NN, which we compare against \textsl{UMAP} which uses the default k-NN. As described in Section 3.1, we tested \textsl{NN}, which connects the nearest neighbor, \textsl{MST-min} which uses the minimum edges from the minimum spanning tree (MST), and \textsl{MST-all} which uses all the edges from the minimum spanning tree. This gives us a connected graph $G'$. Next, we consider two methods for obtaining the local neighborhood for each point, $\mathcal{M}_i$ which is used for the final graph $G$ (Section 3.2). For \textsl{Adjacent Neighbors}, $G'=G$ as there is no change to the local neighborhood. For \textsl{Path Neighbors}, we additionally consider nodes from $G'$ which are more than one hop away until we get $k$ neighbors, to get the `new' local neighborhood  $\mathcal{M}_i$ which is used for $G$.

\paragraph{Hyperparameter Search} For all methods we do a grid search to find the best $k$, the number of initial Nearest Neighbors (before applying the \textit{mutual} NN restriction), by searching from 10 to 50 with increments of 5. We find the best \texttt{min\_dist}, a hyper-parameter that controls how tightly UMAP packs points together, by searching from 0-1 in increments of 0.1. We use Euclidean distance for image datasets, Jaccard for the Count Vectors, and cosine for the TF-IDF Vectors (these distance metrics were best for the original UMAP embeddings).

\section{Results}

From Table 1, we see that \textsl{MST} variants combined with \textsl{Path Neighbors} to find $\mathcal{M}_i$ consistently produced better clustering results across all datasets for both 2 and 64 dimensions ($p < 0.01$).\footnote{Two-tailed t-test for org. UMAP vs Path Neigh. + MST-min.} As a first step to uncover why, we present the 2D projections generated for MNIST, FMNIST, and 20 NG Count Vectors using each method in Fig. 3. We observe that \textsl{MST} variants combined with \textsl{Path Neighbors} consistently produces clearer separation between classes, less “random projections" (better \textit{local structure}), and preserves the “global structure” which leads to consistently better clustering results (Table 1). 

For MNIST, we see that the \textit{global structure} was preserved among different digits, such as having 1 and 0 at far corners and placing similar digits such as 4, 7, 9 closer together. There is also more separation within the groups of similar digits (4, 7 9). Similarly for the FMNIST dataset, the vectors using the aforementioned method preserved the \textit{global structure} between clothing classes(T-shirt, Coat, and etc.) from footwear classes (Sandal, Sneaker, Ankle-boot) while also depicting a clearer separation between the footwear classes. This is contrasted with original UMAP which has poorer separation between similar classes. Finally for 20NG, the generated vectors create a better distinction between similar subjects such as the recreation (rec) topics.

In the following sections, we explore how \textit{mutual} k-NN graph affects separation of classes (5.1), how connectivity reduces random projections (5.2)  and how selecting nearest neighbor can affect the structure of the final vector (5.3).

\begin{figure*}
    \centering
    \includegraphics[width=.8\linewidth]{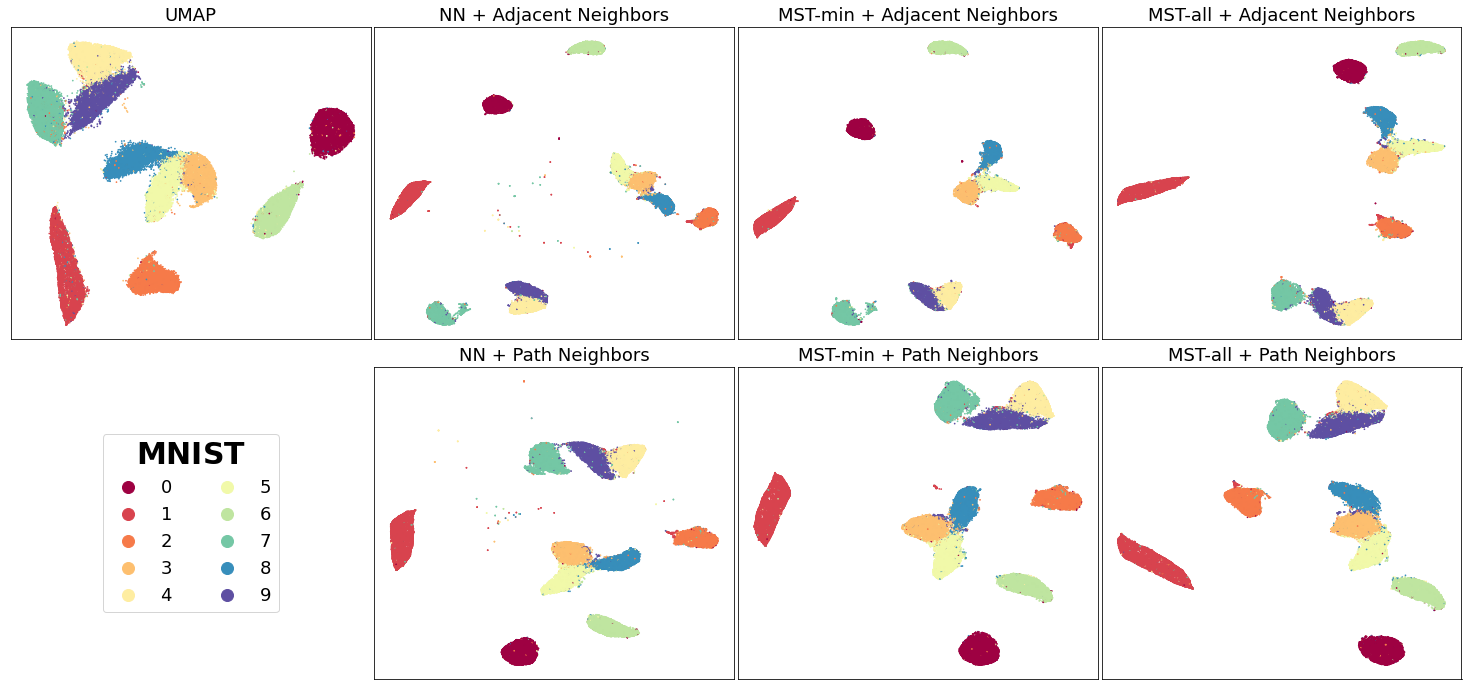}
    \\[\smallskipamount]
    \includegraphics[width=.8\linewidth]{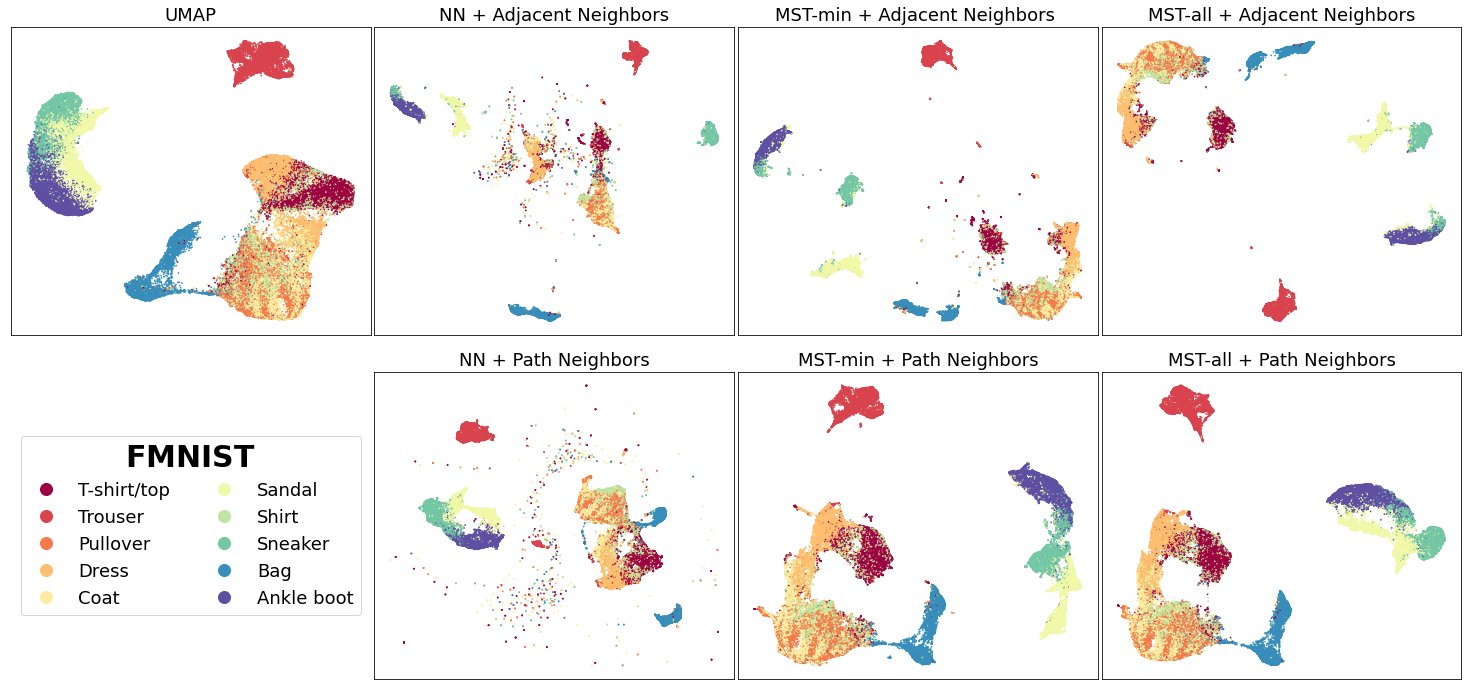}
    \\[\smallskipamount]
    \includegraphics[width=.8\linewidth]{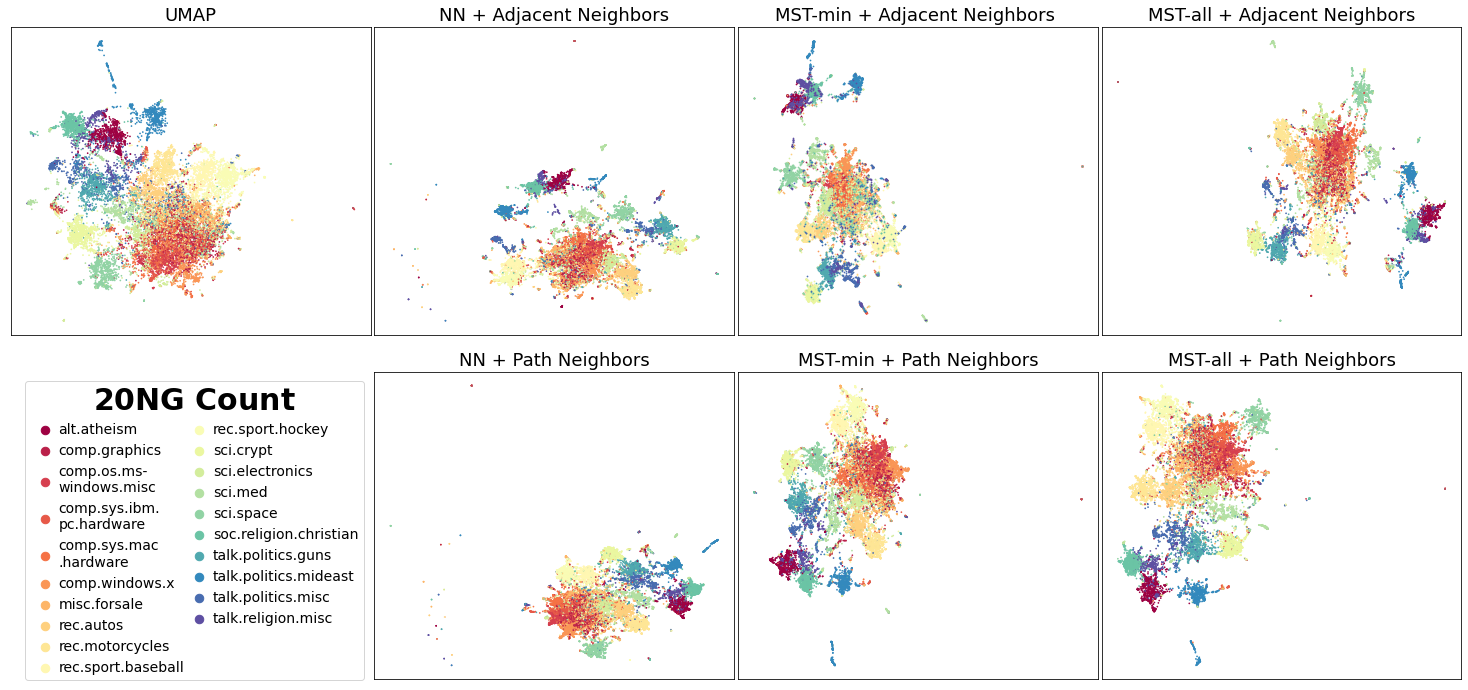}
    \\[\smallskipamount]
    \caption{\small Visual comparisons between all the methods tested in Section 4.3 for MNIST, FMNIST, and 20NG Count. We observe that using the \textbf{MST} variants for Connectivity combined with Path Neighbors improves the separations between locally similar classes while also preserving the global relationships that occur in the datasets.}
    \label{fig:visual}

\end{figure*}


\begin{figure*}[t!]%

    \begin{minipage}{.69\linewidth}

     \centering
     \includegraphics[width=\linewidth, height=3.4cm]{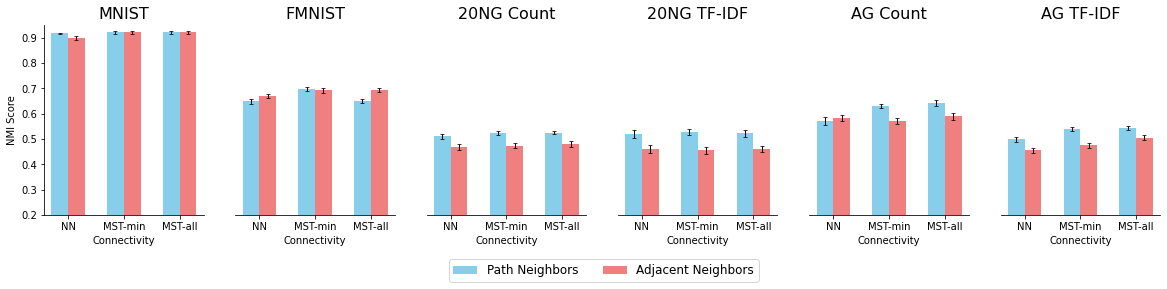}

     \label{fig:NMI}

    \end{minipage}
       \begin{minipage}{.3\linewidth}
    \centering
    \includegraphics[width=0.85\textwidth, height=3.4cm]{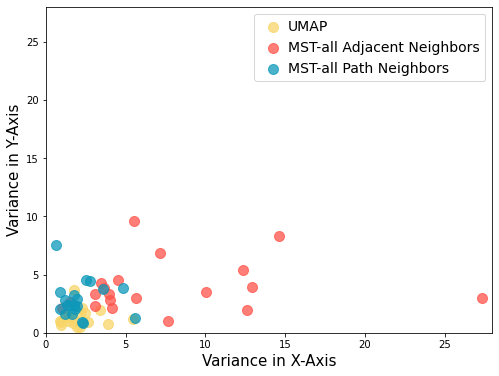} 

    \label{fig:variance}

    \end{minipage}
 \small
     \caption{\small Plots comparing how NMI varied based the Connectivity and Local Neighborhood used. We also plot the variances of 2D vectors generated for each class from the 20NG Count vectors for original UMAP, Adjacent Neighbors and Path Neighbor(using MST-all).}

\end{figure*}

\subsection{Mutual k-NN vs k-NN Graph (default UMAP)}
\label{sec:mutual_knn_r}

\textbf{Using a \textit{mutual} k-NN representation results in improved separation between similiar classes.} In general we observe that for most of the \textit{mutual} k-NN graph based vectors (Fig. 3), there is a better separation between similar classes than the original UMAP vectors regardless of Connectivity (NN, MST variants). 
We observed the desired separation between similar classes such as with the 4, 7, 9 in MNIST and the footwear classes for FMNIST. 
\textit{mutual} k-NN graphs have previously been shown as a useful method for removing edges between points that are only loosely similar, directly reducing \textit{distance concentration} and \textit{hub effects}.


\subsection{Connectivity for \textit{Mutual} k-NN}
We consider three methods for connecting the mutal k-NN. In terms of connectivity, NN $<$ MST-min $<$ MST-all. 

\textbf{NN performs worse than MST variants, with vectors that are randomly scattered in 2d space.} For both MNIST and FMNIST, we see that \textsl{NN}, which connects isolated vertices to their nearest neighbor, had multiple small clusters of points scattered throughout the vector space. Given that KMeans is sensitive to outliers, these randomly projected points negatively affect clustering performance as seen in Table 1. Since a \textit{mutual} k-NN graph only retains a subset of the edges from the original k-NN graph, it can result in a very sparse representation. When constructing the \textit{mutual} k-NN graph, we observed that  $\approx$ 1000($3\%$) points were separated into small components for MNIST and  $\approx$ 7000($10\%$) points for FMNIST. Our results show stronger notions of connectivity than \textsl{NN} are required.

We would expect that having higher connectivity that reduces random scattering of points would be better for clustering. However, we observe that too much connectivity from using all the edges from the MST (MST-all) with Path Neighbors can hurt performance on FMNIST (Section 5.3).

\subsection{Local Neighborhood for Connected \textit{Mutual} k-NN graph}
\label{sec:adj_r}
We consider two methods, Adjacent Neighbors and Path Neighbors, for finding the local neighborhood $\mathcal{M}_i$ of each point, after obtaining a connected \textit{mutual} k-NN graph $G'$.

\textbf{Path Neighbors achieves the best clustering performance together with MST-min.} 
We observe a similar clustering performance when we use the minimum number of edges from the MST (\textsl{MST-min}), vs all the edges from the MST (\textsl{MST-all}). 
 In the case of FMNIST, using MST-all results in a worse clustering performance of 0.698 vs 0.649 for 2 dims, and 0.698 vs 0.679 for 64 dims. Although using \textsl{Adjacent} vs \textsl{Path Neighbors} resulted in similar clustering performance for image based dataset, \textsl{Path Neighbors} produced better results for text based datasets. From Fig. 3, both \textsl{MST-min} and \textsl{MST-all} produced better results when using \textsl{Path} vs \textsl{Adjacent Neighbors} for text based datasets ($p < 0.05$). 

\textbf{Adjacent Neighbors produce a poorer local structure than Path Neighbors.} Visually, the vector generated using the Adjacent Neighbors and MST-min result in disperse dense clusters of points e.g, the footwear classes in FMNIST and the recreation topics in 20 NG. However when we apply \textsl{Path Neighbors}, the groups of points belonging to the same class are less dispersed (Fig. 3). This indicates that Adjacent Neighbors have a poorer local structure than Path Neighbors; i.e. same class  datapoints are far from each other. 



To investigate further why \textsl{Adjacent Neighbors} produce worse results for text datasets, we compute the variance along the dimensions of the 2D vectors for each class from the 20NG Count vectors and plot them in Fig. 4. We find that \textsl{Adjacent Neighbors} has greater variance for each class than the vectors for both \textsl{Path Neighbors} and original \textsl{UMAP}, indicating greater dispersion of points within-class. Having high variance is bad for clustering and indicates a poor lower dimensional representation, as there is no distinct range of values associated with the class label.

\textbf{Path Neighbors increases the connectivity for the final $G$, and therefore can rely on a more refined notion of $G'$ with MST-min instead of MST-all.} We can interpret \textsl{Path Neighbors} as a method which strictly increases the general connectivity of $G$. Consider the local neighborhood $\mathcal{M}_i$ for \textsl{Adjacent Neighbors}, which is not guaranteed to have $k$ connected neighbors for $\mathcal{M}_i$. Points with smaller $\mathcal{M}_i$ will be close to primarily few adjacent neighbors and repelled further away from the other points. This creates small groups of points that belong to the same class being spread across the vector space. On the other hand, for original \textsl{UMAP} and \textsl{Path Neighbor} vectors, $|\mathcal{M}_i|=k$,and local groups of points are more likely to be connected to other groups within the same class. This increase in connectivity explains why visually \textsl{Path Neighbors} methods generate vectors which are less dispersed (within class) than the \textsl{Adjacent Neighbors} method, while still preserving the underlying structure of the data. Across Table 1, we see that using Path Neighbors performs consistently well with MST-min across both image and text datasets, and allows more flexibility in building a locally and globally connected fuzzy simplicial set.




\subsection{Number of Dimensions} We also found that clustering performance was consistent across different dimensions for image datasets but was better at higher dimensions for text dataset with TF-IDF representations. Table 1 show results for 2 and 64 dimensions. While it is not surprising that having 64 dimensions allows the high-dimensional text datasets ($>$26000 features for AG and $>$34000 features for 20NG) to preserve more information, it is interesting that \textsl{MST} variants with \textsl{Path Neighbors} does not produce worse results at 2 vs 64 dimensions.

\section{Conclusion}
The initialization of a weighted connected graph is critical to the success of topology based dimensionality reduction methods. In this work, we established how starting with stricter conditions of connectivity (\textit{mutual} k-NN graph vs standard k-NN graph) results in a better topology. Next, using a flexible method to expand the local neighborhood, and therefore connectivity of the final graph, is best done on a minimally connected (MST-min) mutual k-NN graph. 
Visualisation of the resulting vectors show that they produce better separation between similar classes while preserving both the overall structure of the data. Our quantitative experiments indicate these vectors consistently produce better clustering (empirical gains of relative 5-18\%) across all datasets for both 2 and 64 dimensions despite the simplicity of graph methods, highlighting the role of graph connectivity in topological based dimensionality reduction methods. 

\section*{Acknowledgments}
We thank the anonymous reviewers for helpful feedback. Also Milind Agarwal, Desh Raj, Taryn Wong, and Jinyi Yang for proof-reading and Kelly Marchisio for discussions at an early stage of this project.

\bibliography{aaai22}

\newpage

\appendix

\begin{table*}[tb]
\centering

\label{tab:results_ac}
\tiny
\resizebox{0.8\textwidth}{!}{\begin{tabular}{@{}lll|lll|lll@{}}
\toprule
\multicolumn{3}{l|}{Accuracy} & \multicolumn{3}{l|}{Adjacent Neighbors} & \multicolumn{3}{l}{Path Neighbors} \\  \midrule
Dataset & Dim & UMAP & NN & \begin{tabular}[c]{@{}l@{}} MST-min \end{tabular} & \begin{tabular}[c]{@{}l@{}} MST-all \end{tabular} & NN & \begin{tabular}[c]{@{}l@{}} MST-min \end{tabular} & \begin{tabular}[c]{@{}l@{}}MST-all \end{tabular} \\
MNIST & 2 & 0.852 & 0.966 & \bf{0.968} & \bf{0.968} & 0.929 & \bf{0.968} & \bf{0.968} \\
FMNIST & 2 & 0.578 & 0.670 & 0.665 & \bf{0.692} & 0.610 & \bf{0.693} & 0.549 \\
20NG Count & 2 & 0.489 & 0.524 & 0.520 & 0.531 & \bf{0.562} & \bf{0.570} & 0.558 \\
20NG TF-IDF & 2 & 0.551 & 0.531 & 0.511 & 0.538 & \bf{0.582} & \bf{0.589} & \bf{0.582} \\
AG Count & 2 & 0.802 & 0.831 & 0.823 & 0.833 & 0.827 & \bf{0.859} & \bf{0.865} \\
AG TF-IDF & 2 & 0.805 & 0.751 & 0.755 & 0.795 & 0.780 & \bf{0.842} & \bf{0.841} \\

\midrule
MNIST &  64 & 0.824 & 0.95 & \bf{0.968} & \bf{0.968} & 0.953 & \bf{0.967} & \bf{0.968} \\
FMNIST & 64 & 0.558 & 0.653 & \bf{0.666} & 0.663 & 0.643 & \bf{0.670} & 0.651 \\
20NG Count & 64 & 0.515 & 0.565 & 0.586 & 0.583 & \bf{0.592} & \bf{0.589} & \bf{0.589} \\
20NG TF-IDF & 64 & 0.650 & 0.621 & 0.621 & 0.622 & \bf{0.658} & \bf{0.663} & \bf{0.663} \\
AG Count &64 & 0.810 & 0.854 & 0.857 & 0.857 & \bf{0.869} & \bf{0.869} & \bf{0.873} \\
AG TF-IDF & 64 & 0.810 & 0.804 & 0.802 & 0.809 & 0.840 & \bf{0.842} & \bf{0.847} \\ 
\midrule
\end{tabular}}
\caption{\small Accuracy Results for clustering each of the vectors generated using each method described in Section 4.3 with KMeans.}
\end{table*}

\begin{table*}[tb]
\centering

\label{tab:results_p}
\tiny
\resizebox{0.8\textwidth}{!}{\begin{tabular}{@{}lll|lll|lll@{}}
\toprule
\multicolumn{3}{l|}{Purity} & \multicolumn{3}{l|}{Adjacent Neighbors} & \multicolumn{3}{l}{Path Neighbors} \\  \midrule
Dataset & Dim & UMAP & NN & \begin{tabular}[c]{@{}l@{}} MST-min \end{tabular} & \begin{tabular}[c]{@{}l@{}} MST-all \end{tabular} & NN & \begin{tabular}[c]{@{}l@{}} MST-min \end{tabular} & \begin{tabular}[c]{@{}l@{}}MST-all \end{tabular} \\
MNIST & 2 & 0.888 & 0.966 & \bf{0.968} & \bf{0.968} & 0.939 & \bf{0.968} & \bf{0.968} \\
FMNIST & 2 & 0.639 & 0.700 & 0.707 & \bf{0.721} & 0.677 & \bf{0.724} & 0.645 \\
20NG Count & 2 & 0.508 & 0.538 & 0.535 & 0.547 & \bf{0.576} & \bf{0.583} & \bf{0.579} \\
20NG TF-IDF & 2 & 0.570 &	0.554 &	0.537 &	0.563 &	\bf{0.600} & \bf{0.606} &	\bf{0.606} \\
AG Count & 2 & 0.815 &	0.831 &	0.823 &	0.833 &	0.827&	\bf{0.859} & \bf{0.865} \\
AG TF-IDF & 2 & 0.805 &	0.751 &	0.755 &	0.795 &	0.700 &	\bf{0.829} & \bf{0.826} \\

\midrule
MNIST &  64 & 0.874 & 0.950 & \bf{0.968} & \bf{0.968} & 0.953 & \bf{0.967} & \bf{0.968} \\
FMNIST & 64 & 0.635 & 0.702 & \bf{0.716} & \bf{0.710} & 0.657 & \bf{0.709} & 0.669 \\
20NG Count & 64 & 0.534	& 0.580	& 0.603 &	0.601 &	\bf{0.609} &	\bf{0.607} &\bf{0.607} \\
20NG TF-IDF & 64 & 0.662 &	0.640 &	0.639 &	0.645 &	0.676 &	\bf{0.695} & \bf{0.699} \\
AG Count &64 & 0.820	& 0.854	& 0.854 &	0.857 &	\bf{0.869} & \bf{0.869} &	\bf{0.873} \\
AG TF-IDF & 64 & 0.821	& 0.804	 & 0.802 &	0.809 &	0.840 &	\bf{0.856} & \bf{0.857}
\\ 
\midrule
\end{tabular}}
\caption{\small Purity Results for clustering each of the vectors generated using each method described in Section 4.3 with KMeans.}
\end{table*}

\begin{table*}[tb]
\centering

\label{tab:results_ari}
\tiny
\resizebox{0.8\textwidth}{!}{\begin{tabular}{@{}lll|lll|lll@{}}
\toprule
\multicolumn{3}{l|}{Adjusted Rand Index} & \multicolumn{3}{l|}{Adjacent Neighbors} & \multicolumn{3}{l}{Path Neighbors} \\  \midrule
Dataset & Dim & UMAP & NN & \begin{tabular}[c]{@{}l@{}} MST-min \end{tabular} & \begin{tabular}[c]{@{}l@{}} MST-all \end{tabular} & NN & \begin{tabular}[c]{@{}l@{}} MST-min \end{tabular} & \begin{tabular}[c]{@{}l@{}}MST-all \end{tabular} \\
MNIST & 2  & 0.810 &	0.927 &	\bf{0.931} &\bf{0.931} & 0.885 &	\bf{0.931} & \bf{0.931}\\
FMNIST & 2 & 0.485	& 0.547 &	0.535 &	\bf{0.558} &	0.523 &	\bf{0.580}	& 0.492 \\
20NG Count & 2 & 0.309 & 0.348 & 0.347 &	0.356 &	\bf{0.386} &\bf{0.394} & \bf{0.381} \\
20NG TF-IDF & 2 & 0.398	& 0.351	& 0.330	& 0.349 & \bf{0.403} & \bf{0.421} & \bf{0.413} \\
AG Count & 2 & 0.620 &	0.614 &	0.600 &	0.620 &	0.603 &	\bf{0.669} & \bf{0.669} \\
AG TF-IDF & 2 & 0.530 & 0.464 & 0.498 &	0.540 &	0.519 &	\bf{0.569} & \bf{0.571} \\

\midrule
MNIST &  64 & 0.805 & 0.900 & \bf{0.930} & \bf{0.931} &	0.920 &	\bf{0.930} &	\bf{0.932} \\
FMNIST & 64 & 0.484	& 0.500 & \bf{0.547} & \bf{0.535} & 	0.503 &	\bf{0.539}	& 0.499 \\
20NG Count & 64 & 0.329 &	0.401 &	0.406 &	0.405 &	\bf{0.419} & \bf{0.420} &	\bf{0.420} \\
20NG TF-IDF & 64 & 0.450 & 0.436 &	0.434 &	0.442 &	\bf{0.481} & \bf{0.493} & \bf{0.495} \\
AG Count &64 & 0.660	& 0.662	& 0.662	& 0.668 &	\bf{0.691} &	\bf{0.691} & \bf{0.699}\\
AG TF-IDF & 64 & 0.624	& 0.563	& 0.576 & 0.560 & 0.630 &	\bf{0.665} &	\bf{0.655}\\ 
\midrule
\end{tabular}}
\caption{\small Adjusted Rand Index  Results for clustering each of the vectors generated using each method described in Section 4.3 with KMeans.}
\end{table*}

\end{document}